\documentclass[runningheads]{llncs}
\usepackage{graphicx}
\usepackage{hyperref}

\usepackage{amsmath}
\usepackage{amssymb}
\usepackage{upgreek}
\usepackage{xcolor}
\usepackage{booktabs}
\usepackage{todonotes}
\usepackage{multirow} 
\usepackage{times}
\usepackage{tikz}
\usetikzlibrary{arrows,automata}
\usepackage{url}
\usepackage{marvosym}

\usepackage{longtable}
\usepackage{fancybox,framed}
\usepackage{tabu}
\usepackage{xspace}
\usepackage{bm}
\usepackage{subfig}
\usepackage{caption}

\newcommand{\ours}{\textsc{DeepCDCL}\xspace}

\usepackage{algorithm}  
\usepackage{algorithmicx}
\usepackage[noend]{algpseudocode}

\usepackage{ulem} 
\algrenewcommand{\algorithmiccomment}[1]{\hfill\(\triangleright\) \textcolor[gray]{0.5}{#1}}
\usepackage{multirow}

\begin{document}
\title{\ours: An CDCL-based Neural Network Verification Framework}
%
%
\author{Zongxin Liu\inst{1,2} \and
Pengfei Yang \inst{1} \and
Lijun Zhang\inst{1,2} \and
Xiaowei Huang\inst{3}
}
\authorrunning{F. Author et al.}
%
\institute{SKLCS, Institute of Software, Chinese
Academy of Sciences, Beijing, China \and
University of Chinese Academy of Sciences, Beijing, China \and
The University of Liverpool, Liverpool, United Kingdom\\
}
\maketitle              
\begin{abstract}

Neural networks in safety-critical applications face increasing safety and security concerns due to their susceptibility to little disturbance. In this paper, we propose \ours, a novel neural network verification framework based on the Conflict-Driven Clause Learning (CDCL) algorithm. We introduce an asynchronous clause learning and management structure, reducing redundant time consumption compared to the direct application of the CDCL framework. Furthermore, we also provide a detailed evaluation of the performance of our approach on the ACAS Xu and MNIST datasets, showing that a significant speed-up is achieved in most cases. 
\end{abstract}

\section{Introduction}

With the widespread application of deep neural networks (DNNs) in safety-critical areas such as flight control~\cite{julian2019deep} and autonomous driving~\cite{selfdriving}, any mistake in their decision-making process can lead to catastrophic consequences. 
Ensuring the safety of neural networks has become increasingly important. 
Formal verification is a technique to prove the correctness of hardware or software systems. 
Verification of neural networks can be time-consuming, which limits its practical application. To facilitate its broader application, it is necessary to explore more efficient DNN verification frameworks.


Most state-of-the-art DNN verification tools are based on the basic branch-and-bound framework. These tools primarily focus on accelerating and improving the accuracy of computing the bounds for each neuron, as well as selecting the branching order.
For instance, $\alpha,\beta$-CROWN~\cite{xu2020fast,wang2021beta,zhang2022general} employs GPU acceleration to compute neuron bounds and uses Lagrange multipliers to encode the constraints of branching decisions, thus enhancing the accuracy of the bounds. MN-BaB~\cite{ferrari2022complete} utilizes Lagrange multipliers to encode multi-neuron constraints and improve bound accuracy. Marabou~\cite{marabou} uses SOI local search to determine a suitable branching order, with the decision based on the neuron that has the most significant impact on the output layer. LayerSAR~\cite{layerSAR} and PeregriNN~\cite{khedr2021peregrinn} select the neuron closest to the input layer for branching.


However, these methods do not fully utilize infeasible paths generated during verification, which could be used to prune the search space.
To address this limitation, some verification frameworks have been introduced. For example, Planet~\cite{ehlers2017formal} incorporated the SAT framework into neural network verification. However, Planet did not fully exploit the impact of node branching on boundaries. Additionally, the selection of split nodes in Planet is determined by the SAT solver at the SAT problem level, without fully leveraging neural network characteristics.
Moreover, the traditional SAT framework is not well suited for parallel solving and asynchronous learning of conflict clauses, leading to additional time consumption in the solving process. This highlights the need for further research and development of verification tools that can effectively leverage infeasible paths and optimize the verification process for neural networks.

Based on the aforementioned circumstances, we propose the \ours framework. This framework is built upon the Conflict-Driven Clause Learning (CDCL) algorithm, which leverages conflict clauses to prune the search space and accelerate the solving process. In order to further enhance the efficiency of the CDCL framework, we introduce an asynchronous clause learning and management structure that mitigates additional time consumption during the solving process. 

The contributions of this work can be succinctly outlined as follows.

\begin{enumerate}
    \item We implement \ours, a novel neural network verification framework integrating state-of-the-art neural network verification tools with the CDCL framework.
    \item To address the characteristics of neural network verification problems, we introduce an asynchronous clause learning and management structure, reducing redundant time consumption compared to the direct application of the CDCL framework.
    \item We demonstrate specific speed-ups of \ours on two datasets and provide an illustration of the acceleration of \ours in branch cutting. 
\end{enumerate}

\section{Background}



The Neural Network Verification Problem is defined as a tuple $\langle f, \mathcal{C}, \mathcal{P}\rangle$, aiming to determine whether for all $x \in \mathcal C$, it holds that $f(\boldsymbol{x}) \in \mathcal P$, where $f$ is a neural network, $\mathcal{C}$ is the input constraint, and $\mathcal{P}$ is the property to be verified. 
If an input $\boldsymbol{x}^*\in\mathcal{C}$ satisfies $f(\boldsymbol{x}^*)\not\in \mathcal{P}$, we call $\boldsymbol{x}^*$  a counterexample. In this paper, we focus on neural networks with ReLU activation functions, which is defined as $\mathrm{ReLU}(x)=\max(0,x)$ for $x \in \mathbb R$.


A neural network verification procedure typically involves three components: constraint solving, bounding computation, and branch selection, where the latter two are known as the Branch-and-Bound (BaB) process. It begins with abstracting each neuron using linear relaxations~\cite{deeppoly} and calculating its upper/lower bound. If property verification cannot be achieved through the bounding computation, more accurate constraint solving like linear programming (LP) is employed. If constraint solving still cannot  determine the property validity, the process proceeds to branch selection on ReLU activations, where a ReLU relation with uncertain activation patterns is split into the activated branch and the deactivated one.

Conflict-Driven Clause Learning (CDCL)~\cite{marques2021conflict} is a vital technique widely used in contemporary SAT and SMT solvers. The CDCL framework integrates conflict clause learning, unit propagation and other techniques to improve the efficiency, showing its effectiveness in many branch-and-bound problems like SAT and linear SMT~\cite{DBLP:journals/jsat/Sebastiani07,DBLP:journals/jacm/NieuwenhuisOT06}, but it has not been well suited for neural network verification. 

\section{\ours Verification Framework}

In this section, we present \ours, a novel CDCL-based neural network verification framework.

\subsection{Overview of \ours}

\begin{figure}[t]
  \centering
  \includegraphics[page=8, trim={110mm 160mm 110mm 52mm}, clip, width=\linewidth]{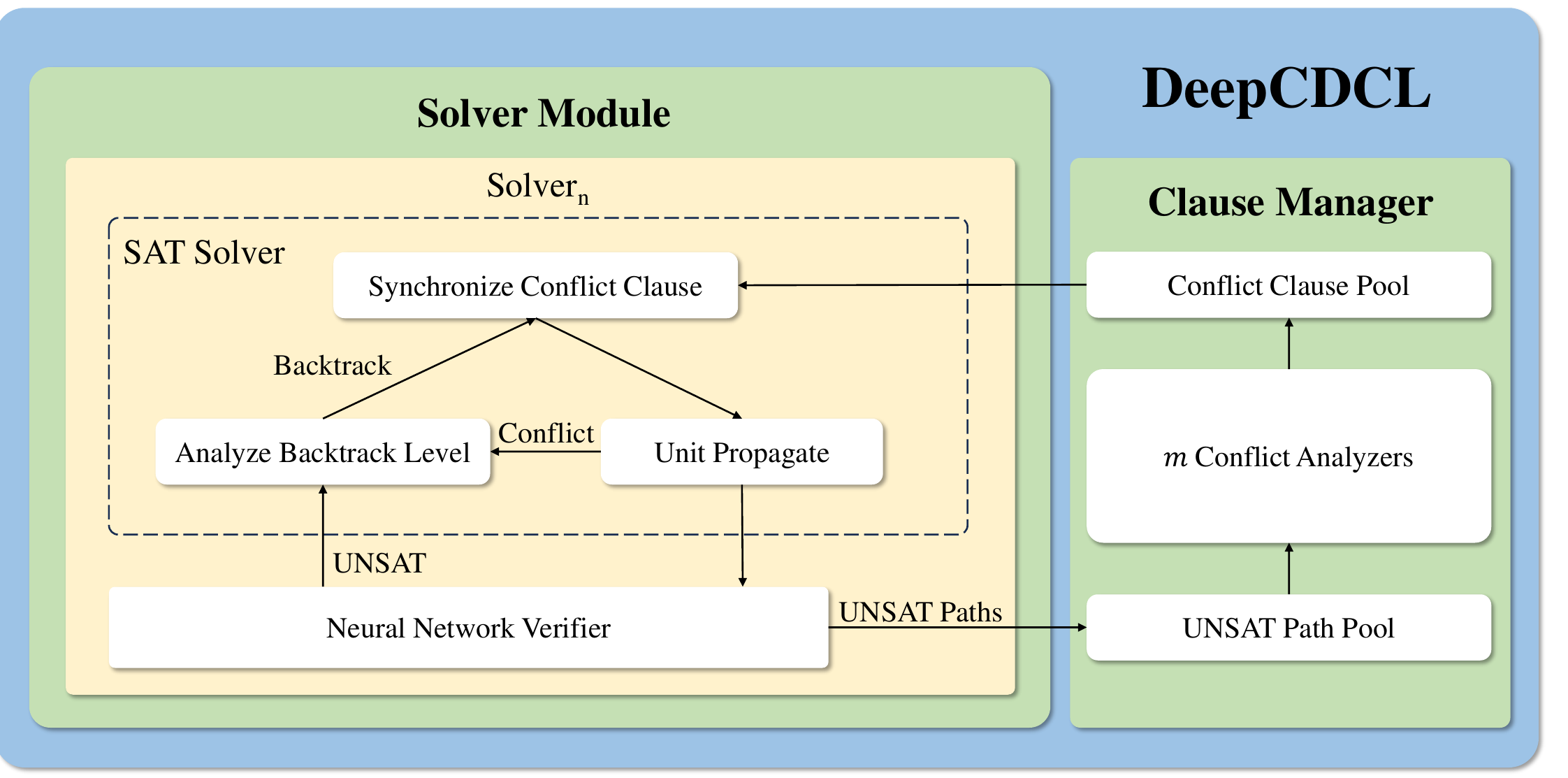}
  \caption{\ours Framework}
  \label{fig: DeepCDCL_framework}
\end{figure}

The comprehensive workflow of the \ours framework is illustrated in Fig.~\ref{fig: DeepCDCL_framework}~, which outlines two pivotal modules: the Solver module and the Clause Manager. The Solver module undertakes the task of problem resolution by installing a set of $n$ Solver threads. Each thread employs an SAT solver to apply conflict clauses that have been collected earlier and call a neural network verifier to solve subproblems.
In contrast, the Clause Manager module generates, collects and manages conflict clauses. This involves a collection of $m$ Conflict Analyzer threads specifically designed to analyze unsatisfiable paths during the Solving Process. The Clause Manager maintains two distinct pools: the Conflict Clause Pool and the UNSAT Path Pool. These pools serve as repositories for conflict clauses generated by the $m$ Conflict Analyzers and unsatisfiable paths generated by the $n$ solvers, respectively.

At the beginning of the verification process, we initiate $m$ Solver threads for solving and concurrently activate $n$ Conflict Analyzers threads to analyze unsatisfiable paths in the Solver's solving process, trying to find conflict clauses. Each solver attempts to retrieve conflict clauses from the Clause Pool to synchronize all newly added conflict clauses at the beginning of each verification loop.
Subsequently, the solver initiates the unit propagation process. 

If a conflict arises during the unit propagation, a conflict analysis is conducted to determine the backtracking level, and the solver backtracks accordingly. Otherwise, the solver verifies the abstract state of the current neural network. If the current search path is determined to be unsatisfiable through bounding computation or constraint solving, the unsatisfiable path is submitted to the UNSAT Path Pool, followed by an analysis of the backtracking level and subsequent backtracking. If backtracking is unfeasible, it signifies the unsatisfiability of the overall problem, enabling the affirmation that the property holds. If the constraints are determined to be satisfiable through bounding computation or constraint solving, the solver initiates a localized search for a counterexample. If a true counterexample is found, the property is directly asserted to be violated; otherwise, branching occurs, initiating the next verification loop.

\subsection{Implementation and Optimization}

In our implementation, we use Z3~\cite{conf/tacas/MouraB08} as the SAT solver and Marabou~\cite{marabou}, the second-place tool in VNN-COMP 23~\cite{brix2023vnn},
as the neural network verifier. We have developed three conflict clause generation methods tailored for neural network verification and use Gurobi~\cite{gurobi} as the LP solver to solve the constraints.

During the search process, if a path is determined to be unsatisfiable, we incorporate it as a conflict clause by adding the negation of the disjunction of the current search path to the original problem's encoding.
Following each branching, we propagate the bounds. If the lower bound of a neuron exceeds 0 or the upper bound falls to 0 during this process, we can determine its activation state and encode this as learned clause.
Drawing inspiration from~\cite{ehlers2017formal}, we utilize elastic filtering~\cite{Chinneck1991LocatingMI} to identify conflict clauses. This method introduces slack variables to the constraints corresponding to each branching decision in an unsatisfiable search path, based on the initial constraints. The sum of these slack variables acts as an optimization objective. By selecting the slack variable(s) with the maximum relaxation and setting them to 0 until the constraints become unsatisfiable, we identify the constraints associated with slack variables set to 0 as the conflict clauses. This approach prioritizes constraints requiring greater relaxation, which are more likely to lead to conflicts. 

However, we find that elastic filtering often consumes a significant amount of time. Therefore, we have transformed it into an asynchronous process. We use a conflict clause pool to store the learned conflict clauses so that a clause generated from one solver can be used by other solvers. We also add a timestamp to each unsatisfiable path and only process the latest path when learning conflict clauses. This approach reduces redundant time consumption and ensures that the learned clauses are up-to-date.
What's more, when the constraints cannot be determined to be unsatisfiable even after adding all constraints, traditional elastic filtering will waste too much time. To address this issue, we introduce a binary search method to find conflict clauses based on the original algorithm. This method effectively filters out undecidable cases by adding all constraints at the beginning, accelerating the process of finding conflict clauses.

Additionally, we also propose several optimizations. We observed that partitioning the input space a limited number of times can significantly accelerate the verification especially when the number of inputs is small. Therefore, we set a threshold to stop the input partitioning when the number of input partitions exceeds the threshold. We also streamlined the complex and redundant data structures in Marabou during backtracking, retaining only the core components necessary for verification. To expedite the learning of conflict clauses and ensure the discovery of conflict clauses, we used a separate thread for learning and synchronized the results in each verification loop.

\section{Evaluation}

In our experimental set-up, we utilized a compute server equipped with 256 AMD EPYC 7702 64-core processors, and 2TB of memory, and operated on the GNU/Linux operating system. The Marabou tool served as our primary framework, providing  specifications to carry out our experiments.



\begin{figure}[t]
  \centering
  \begin{minipage}{.55\textwidth}
    \centering
    \includegraphics[width=0.49\textwidth]{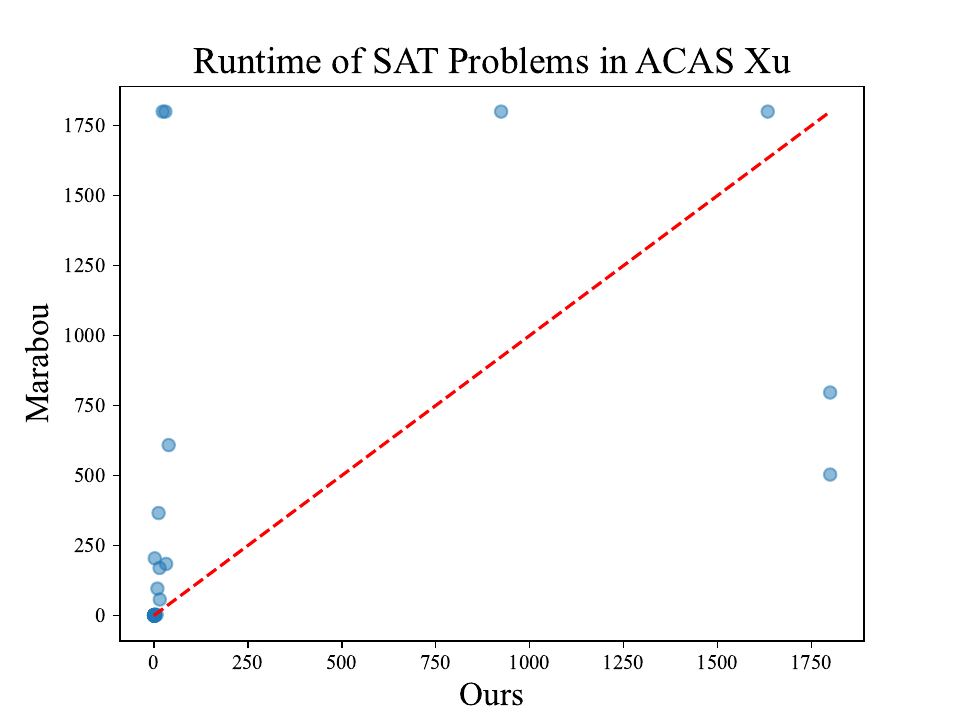}
    \includegraphics[width=0.49\textwidth]{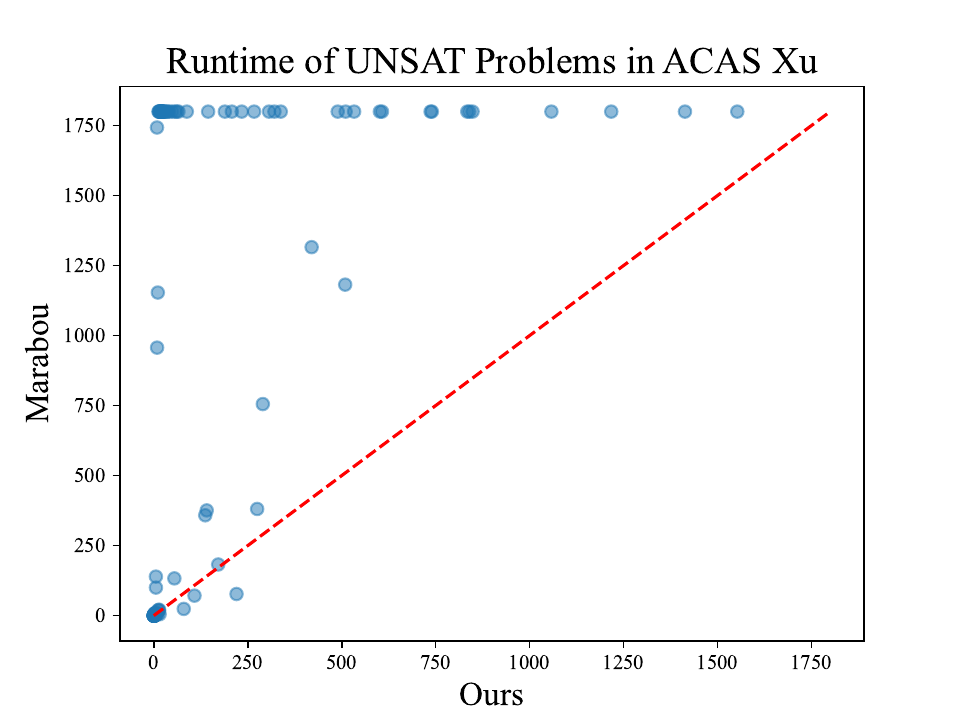}
    \includegraphics[width=0.49\textwidth]{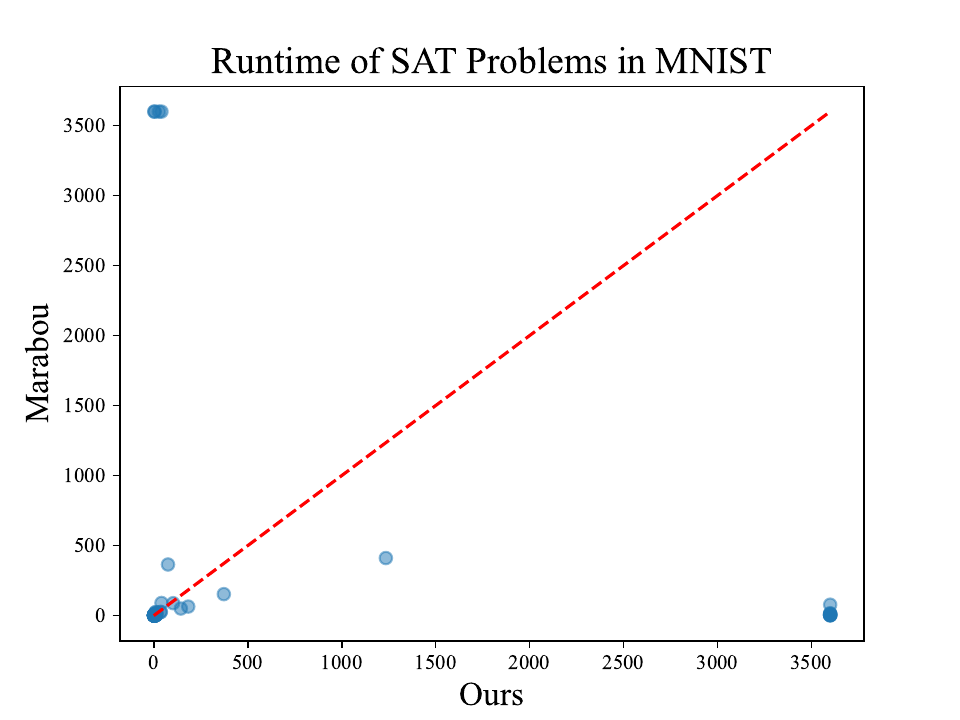}
    \includegraphics[width=0.49\textwidth]{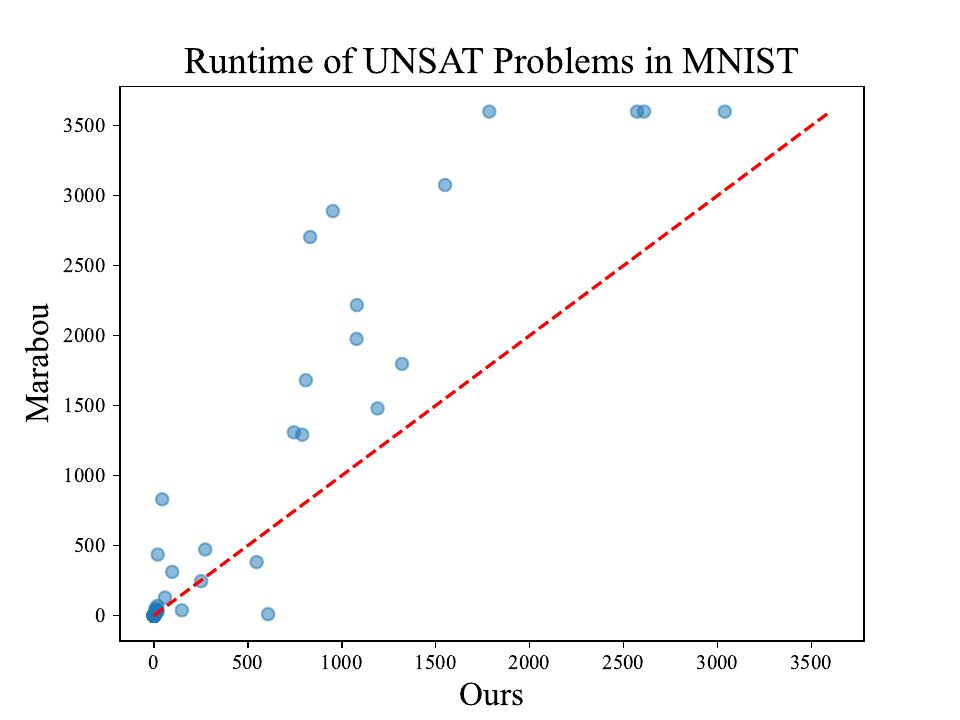}
    \captionof{figure}{Comparison of the Running time}
    \label{fig: runtime compares}
  \end{minipage}%
  \hfill
  \begin{minipage}{.43\textwidth}
    \centering
    \includegraphics[page=4, clip, trim=180 60 180 60, width=\textwidth]{Img/case_study/case_study.pdf}
    \captionof{figure}{Case study: search trees (all nodes) generated by Marabou and \ours(colored nodes)}
    \label{fig: case study}
  \end{minipage}
\end{figure}

For ACAS Xu~\cite{julian2016policy}, which focuses on the safety of aircraft collision avoidance systems, we employed neural networks consisting of 6 layers with 50 neurons in each layer. In our analysis a total of 45 neural networks were utilized, accompanied by four specifications of the Marabou tool. To ensure efficient execution, we established a timeout of 1800 seconds for each verification task.
In the case of the MNIST dataset~\cite{deng2012mnist}, designed for recognizing handwritten digits, seven neural networks with varying sizes, namely $2\times 256$, $4\times 256$, $6\times 256$, $10\times 10$, $10\times 20$, $20\times 20$, and $20\times 40$, were employed. Specifications were crafted by selecting three images, setting robustness radii at 0.01, 0.03, and 0.1, with classification results ranging from 0 to 9. We chose 81 specifications from the initial set of specifications, thus forming a total of 567 verification tasks. We then exclude 97 tasks that can be successfully attacked by the PGD attack~\cite{pgd}. A timeout of 3\,600 seconds was imposed on each verification task.
All the experiments were executed with 20 concurrent threads and we set $n=10$ and $m=10$ for \ours.

Fig.~\ref{fig: runtime compares}~shows the comparison of the running time of \ours and Marabou. 
Overall, \ours outperforms Marabou in most cases, solving problems faster than Marabou. 
Specifically, our approach successfully resolves 2 additional problems outputting SAT and 47 more UNSAT problems compared to Marabou. Furthermore, our method achieves notable computational efficiency, solving 1 problem 205 times faster than Marabou, while 11 problems exhibit over 100 times speedup and 24 problems demonstrate more than a 10-fold acceleration relative to Marabou.

On MNIST, \ours exhibits superior performance in terms of speed for almost all the UNSAT problems compared to Marabou. Our approach resolves 4 additional UNSAT problems, with one instance achieving a remarkable 1\,478 times speedup. Additionally, 75 problems demonstrate a minimum of 25\% acceleration.
However, \ours does not perform so well in the problems outputting SAT, solving 6 fewer than Marabou, but still solving 3 problems that the Marabou tool cannot solve.

To further illustrate the reasons for our acceleration on the UNSAT problems, we present a case study in Fig.~\ref{fig: case study}. The figure shows the search trees of a $2\times 256$ neural network with a specified image 1, target 6 and the $L_\infty$ robustness radius 0.05. The whole search tree is generated by Marabou and the sub-tree of colored nodes is generated by \ours with red indicating the conflict clause learned from the corresponding search path.
For simplicity, we use literal $a$ to represent that a neuron is activated.
Specifically, at node 3, a conflict clause $a$ is identified and at node 5, the conflict clause $\lnot a \wedge b$ is retrieved. Consequently, it is deduced that $\lnot b$ must always be true. Subsequently, a backtrack to node 1 occurs, where the constraint $\lnot b$ is added to the solver. This action renders certain constraints unsatisfiable for Marabou, implying that $c$ cannot be true.
At node 8, the conflict clause $\lnot a \wedge \lnot c$ is retrieved, leading to the direct conclusion that the problem is unsatisfiable. Compared to Marabou, which explores 179 states and 90 UNSAT paths, \ours only explores 17 states and 5 UNSAT paths. Reduced exploration signifies less time consumption, elucidating the reason for our superior speed on the UNSAT problems.

However, in the problems outputting SAT, Marabou can leverage multiple threads to explore a broader search space. In contrast, \ours, when venturing into a sub-tree where the problem is unsatisfiable, persists until proving the unsatisfiability of the sub-problem. This determination may result in an increase in time consumption.

\section{Related work}

The neural network verification problem was first proposed in 2010 \cite{pulina2010abstraction}.
Since 2017, the field of neural networks verification has rapidly developed, with many neural network verification tools and various verification methods emerging, such as abstract interpretation \cite{gehr2018ai2,mirman2018differentiable,singh2018fast,singh2018boosting,deeppoly,singh2019beyond}, symbolic propagation \cite{wang2018formal,weng2018towards,wang2018efficient,gowal2019scalable,boopathy2019cnn,li2019analyzing,zhang2019towards}, SMT \cite{katz2017Reluplex,ehlers2017formal,huang2017safety,gopinath2017deepsafe,katz2019marabou,bastani2016measuring}, convex optimization \cite{muller2022prima,anderson2019optimization}, mathematical programming \cite{bertsimas1997introduction,tran2020nnv,tjeng2017evaluating}, counterexample-guided abstraction refinement \cite{anderson2019optimization,yang2021improving,ashok2020deepabstract}, and Lipschitz constants \cite{weng2018evaluating,weng2018towards}.

\section{Conclusion}
In this paper, we propose a novel neural network verification framework, \ours, which is based on the CDCL framework and tailored to the context of neural network verification. We show that \ours leads to faster verification. We provide a detailed evaluation of \ours on the ACAS Xu and MNIST datasets, showing that \ours is more efficient than Marabou in most cases, especially for unsatisfiable verification problems. We also provide a case study to illustrate why \ours outperforms in solving UNSAT problems. In the future, we plan to further optimize the \ours framework by designing heuristic CDCL strategies and incorporate our framework into other advanced tools.
\bibliographystyle{splncs04}
\bibliography{ref}

\end{document}